\documentclass[runningheads]{llncs}
\usepackage{color}
\usepackage{url}
\usepackage[belowskip=-12pt,aboveskip=-12pt]{caption}
\usepackage{amssymb,amsmath,amsfonts}
\usepackage{array}
\usepackage{tikz}
\usepackage{pgfplots}
\usepackage{pgfplotstable}
\usepackage{multicol}
\usepackage{graphicx}
\usepackage{tabularx}
\usepackage{multirow}
\usepackage{rotating}
\usepackage{hyperref}
\usepackage{array}
\usepackage{caption}
\usepackage{subfig}
\usepackage{xspace}
\usepackage{xifthen}
\usepackage[notheorems]{funkey}

\usetikzlibrary{arrows}
\usetikzlibrary{backgrounds}
\usetikzlibrary{calc}
\usetikzlibrary{chains}
\usetikzlibrary{fit}
\usetikzlibrary{matrix}
\usetikzlibrary{mindmap}
\usetikzlibrary{pgfplots.colormaps}
\usetikzlibrary{positioning}
\usetikzlibrary{shadows}
\usetikzlibrary{shapes}
\usetikzlibrary{trees}
\usetikzlibrary{3d}

\usetikzlibrary{arrows}
\usetikzlibrary{backgrounds}
\usetikzlibrary{calc}
\usetikzlibrary{chains}
\usetikzlibrary{fit}
\usetikzlibrary{matrix}
\usetikzlibrary{mindmap}
\usetikzlibrary{pgfplots.colormaps}
\usetikzlibrary{positioning}
\usetikzlibrary{shadows}
\usetikzlibrary{shapes}
\usetikzlibrary{trees}
\usetikzlibrary{fit}
\usetikzlibrary{3d}

\tikzstyle{helper}=
  [inner sep=0,outer sep=0]

\tikzstyle{supervoxel}=
  [circle,fill=white,draw=black]
\tikzstyle{merge}=
  [rectangle,rounded corners,draw]

\tikzstyle{adjacency}=
  [draw=gray]
\tikzstyle{subset}=
  [draw=black,thick,->]

\tikzstyle{candidate}=
  [circle,fill=gray!20!white,draw=black]

\tikzstyle{inactive}=
  [draw opacity=0.1,fill opacity=0.1]

\tikzstyle{annotation}=
  [rectangle,rounded corners,fill=orange!10!white,draw=orange!10!black]

\tikzstyle{task}=
  [rectangle,rounded corners,fill=blue!50!white]

\pgfplotsset{compat=1.12}

\pgfplotsset{
  errors/.style={
    stack plots=y,
    area style,
    enlarge x limits=false,
    xmajorgrids=true,
    ymajorgrids=true,
    yminorgrids=true,
    legend reversed
  }
}

\pgfplotsset{compat=1.9}

\def\scale#1#2{\tikz[scale=#1,transform shape]{\node{#2};}}

\def\reffig#1{Fig.~\ref{#1}\xspace}
\def\reftab#1{Table~\ref{#1}\xspace}
\def\refsec#1{Section~\ref{#1}\xspace}
\def\refeq#1{Eq.~\ref{#1}\xspace}

\def\ie{i.e.\xspace}

\begin{document}
\title{The Candidate Multi-Cut for Cell Segmentation}
\titlerunning{Candidate Multi-Cut}

\author{Jan Funke$^1$, Chong Zhang$^2$, Tobias Pietzsch$^3$, Stephan Saalfeld$^4$}
\authorrunning{Funke et\,al.}
\institute{
  $^1$ Institut de Robotica i Informatica Industrial, UPC, Barcelona \\
  $^2$ Universitat Pompeu Fabra, Barcelona \\
  $^3$ MPI-CBG, Dresden \\
  $^4$ Janelia Research Campus, Ashburn VA
}

\maketitle

\begin{abstract}

Two successful approaches for the segmentation of biomedical images are (1) the selection of segment candidates from a merge-tree, and (2) the clustering of small superpixels by solving a Multi-Cut problem. In this paper, we introduce a model that unifies both approaches. Our model, the \emph{Candidate Multi-Cut} (CMC), allows joint selection \emph{and} clustering of segment candidates from a merge-tree. This way, we overcome the respective limitations of the individual methods: (1) the space of possible segmentations is not constrained to candidates of a merge-tree, and (2) the decision for clustering can be made on candidates larger than superpixels, using features over larger contexts. We solve the optimization problem of selecting and clustering of candidates using an integer linear program. On datasets of 2D light microscopy of cell populations and 3D electron microscopy of neurons, we show that our method generalizes well and generates more accurate segmentations than merge-tree or Multi-Cut methods alone.

\end{abstract}

\section{Introduction}
\label{sec:introduction}

\begin{figure}[t]
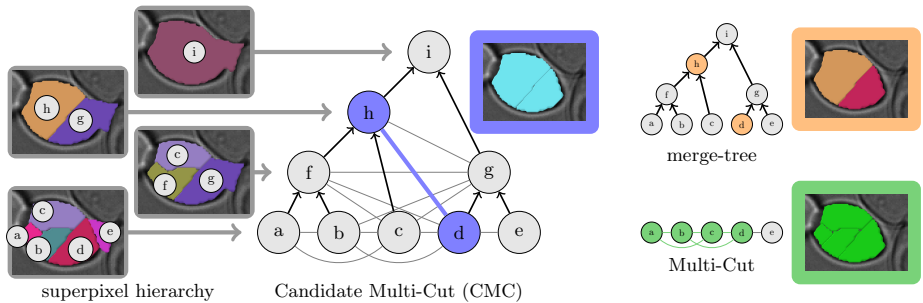

  \includetikz[scale=0.8]{figures/cmc/comparison/comparison}
  \vspace{5mm}
  \caption[]{Illustration of CMC compared to a merge-tree and Multi-Cut segmentation. In the superpixel hierarchy shown on the left, there is no candidate producing the desired segmentation of the cell. Consequently, the best merge-tree segmentation (which does not use the adjacency edges between candidates) is the selection of candidates $d$ and $h$ as two separate objects (highlighted in orange) which introduces a split error. CMC, in contrast, introduces adjacency edges between all neighboring candidates. Therefore, CMC can pick candidates $d$ and $h$, \emph{and} the adjacency edge $(d,h)$ to merge these candidates (highlighted in blue). This way, a segment can be assembled that is not limited by the heuristic superpixel hierarchy. The Multi-Cut on the same initial superpixels could possibly achieve the same segmentation, but has to resort to smaller superpixels that lack structural information (highlighted in green).}
  \label{fig:cmc:comparison}
  \vspace{-2mm}
\end{figure}

\begin{figure}[t]
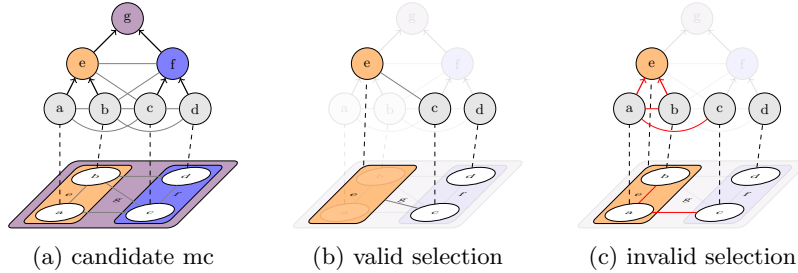

  \hspace{0.1cm}
  \subfloat[candidate mc]{
    \includetikz[scale=0.6]{figures/cmc/overview}
    \label{fig:cmc:overview:cmc}
  }
  \hspace{0.2cm}
  \subfloat[valid selection]{
    \includetikz[scale=0.6,define=showvalid]{figures/cmc/valid}
    \label{fig:cmc:overview:valid}
  }
  \hspace{0.2cm}
  \subfloat[invalid selection]{
    \includetikz[scale=0.6,define=showinvalid]{figures/cmc/invalid}
    \label{fig:cmc:overview:invalid}
  }
  \vspace{5mm}
  \caption[]{Illustration of the proposed model. 
  \subref{fig:cmc:overview:cmc} Lower part: Given initial supervoxels 
  ($a$,$b$,$c$, and $d$), a heuristic region merging strategy is used to generate 
  larger candidates ($e$,$f$, and $g$). Upper part: The \emph{candidate region 
  adjacency graph} (CRAG) is used to represent all candidates, their adjacencies across \emph{all} levels (solid gray lines), 
  and subset relations (black arrows). \subref{fig:cmc:overview:valid} A 
  selection of candidates and adjacency edges producing a valid segmentation. Note that this segmentation can not be produced by the merge-tree method alone, since it can not merge $e$ and $c$. A Multi-Cut method could generate the same segmentation by merging $a$, $b$, and $c$, but would not be able to exploit the features extracted on the larger candidate $e$.
  \subref{fig:cmc:overview:invalid} An invalid selection of candidates 
  and adjacency edges: Candidates $a$, $b$ overlap with $e$ and can not be selected 
  with $e$ at the same time. Further, $a$ is merged both with $b$ and $c$, but 
  $b$ and $c$ are not merged, thus violating the transitivity of equivalence.}
  \vspace{-2mm}
\end{figure}

%
In this paper, we are addressing the problem of segmenting multiple objects in biomedical images, possibly against background. For this problem, \emph{merge-tree methods} and \emph{Multi-Cut methods} are amongst the best performing for a range of data modalities.

  Both methods start with an initial set of superpixels, that is assumed to provide an oversegmentation.
  In {\bf merge-tree methods}, these superpixels are iteratively merged to obtain a hierarchy of candidate segments.
  Amongst all candidates, a cost-minimal and non-overlapping subset is selected to yield a segmentation.
  The advantage of these methods is that they can consider candidates larger than the initial superpixels and thus use more meaningful features.
  A clear disadvantage is the limited expressiveness, as these methods require that each object is correctly segmented by one candidate in the merge-tree.
  Merge-tree methods demonstrate state-of-the-art performance for the segmentation of cells in 2D light microscopy~\cite{Arteta2013,Arteta2016,Funke2015a}.
  Variations of this approach have also been successfully applied to the segmentation of neurons in volumes of electron microscopy~\cite{Funke2012,Kaynig2015}, but were ultimately outperformed by Multi-Cut methods~\cite{snemi_2016_03}.
  {\bf Multi-Cut methods} consider finding a segmentation as an instance of a clustering problem on superpixels~\cite{Andres2012,Kroeger2013}.
  For that, edges in an adjacency graph of superpixels are cut.
  A segmentation is obtained as the connected components of a cost-minimal cut, where constraints ensure that there is no path connecting two separated superpixels.
  In contrast to merge-tree methods, a correct segmentation can theoretically always be obtained, if the initial superpixels are oversegmenting.
  However, small superpixels carry the risk of not capturing meaningful features, like the local orientation of a cellular structure or the diameter of a cell, which could help to resolve ambiguities during inference.
  Multi-Cut methods are the current state of the art for the segmentation of neurons in electron microscopy volumes~\cite{Beier2015,Kroeger2013,Andres2012,snemi_2016_03}, but are outperformed by merge-tree methods on the segmentation of cells in light microscopy images~\cite{Funke2015a}.

The specific advantages and disadvantages of the two methods make them perform differently depending on the characteristics of a given dataset.
  In particular, none of the two methods performs well on \emph{both} 2D segmentation of cells in light microscopy and 3D segmentation of neurons in electron microscopy.

%
%
To combine the advantages of both methods (larger feature context of merge-tree methods and the expressiveness of Multi-Cut methods), we introduce a segmentation model that jointly selects \emph{and} clusters segment candidates from a merge tree.
  First, we obtain a merge-tree of segment candidates following greedy merging on initial superpixels.
  We then introduce adjacency edges between all adjacent candidates across all levels of the tree and
  train a classifier on ground-truth to obtain a cost for the selection of each candidate and each merge of an adjacency edge.
  Finally, we find the globally cost-minimal selection and clustering of candidates by solving an integer linear program (ILP) with a cutting-plane method.
%
%
Compared to the standard formulations for merge-tree and Multi-Cut segmentations, our model has two advantages:

First, our model unifies the two method families including both as special cases (see \reffig{fig:cmc:comparison}).
  A merge-tree segmentation in the style of~\cite{Arteta2013,Funke2015a} can be obtained by simply omitting the adjacency edges.
  Similarly, the Multi-Cut formulation~\cite{Andres2012,Kroeger2013} can be obtained by omitting candidates other than the initial superpixels.
  In our model, however, a valid solution allows to select higher-level candidates and merge them with lower-level candidates.
  This allows us to train a classifier on more meaningful features that are only available for larger candidates.
  In contrast to merge-tree methods, however, we are not limited by the choice of the extracted candidates.  Every possible segmentation given the initial superpixels can still be realized.

Second, by allowing candidates to not be selected at all, our formulation is particularly well suited to segment several foreground objects against background.
  This is required in 2D cell segmentation from light microscopy images where foreground objects are not tiling the plane.
  Our model has a dedicated cost contribution for the selection of candidates which depends on features of foreground objects.
  This is in contrast to previous Multi-Cut methods that required a post-processing step to filter background segments from foreground segments~\cite{Zhang2014,Yarkony2015}.

%
We evaluate our model in two different and dissimilar setups: First, on the segmentation of cells in 2D light microscopy (involving three datasets of different resolution, microscopy modality, and cell types), and second, on the reconstruction of neurons from electron microscopy volumes of neural tissue. Our model shows a consistent improvement over both merge-tree methods and Multi-Cut methods. This is of particular interest since neither merge-tree methods nor Multi-Cut methods deliver state-of-the-art performance on \emph{both} datasets jointly.

\section{Method}
\label{sec:method}

In order to combine the advantages of merge-tree methods and Multi-Cut
methods, we introduce a model generalizing both: the \emph{Candidate Multi-Cut} (CMC).
  In our model, the standard Multi-Cut formulation~\cite{Andres2012} is extended 
  by considering additional candidate segments formed by merging initial
  superpixels to obtain a merge-tree. Various merging strategies can be used to obtain a merge-tree. In \refsec{sec:results}, we show two strategies for the datasets used here.

Given a merge-tree, we are addressing a segmentation problem in terms of the selection of candidate regions and adjacency edges (see \reffig{fig:cmc:overview:cmc}). Let $G=(V,E,S,f,g)$ be the
\emph{candidate region adjacency graph} (CRAG), where $V$ is the set of all
candidate regions (including the original superpixels), $E \subset V\times V$
the set of undirected edges indicating region adjacencies across all levels of the merge-tree, $S \subset V\times V$ the set of directed
edges indicating subset relations of the candidate regions of the merge-tree, and $f: V \mapsto
\mathbb{R}$ and $g: E \mapsto \mathbb{R}$ are cost functions for the selection
of candidates and adjacency edges to merge, respectively. These costs are trained on features extracted for candidates and adjacency edges, see \refsec{sec:results} for details. We encode a selection
and merging of candidates with binary indicator variables $\vct{y} = ( y_i \in
\{0,1\} \;|\; i \in V)$ and $\vct{m} = (m_e \in \{0,1\} \;|\; e \in E)$.
Setting $y_i = 1$ means that the candidate represented by node $i$ is part of an
object (as opposed to being considered background). Setting $m_{(i,j)} = 1$
states that the adjacent candidates $i$ and $j$ are part of the same object.
By rewriting the costs $f(i)$ and $g(e)$ as vectors $\vct{f} = (f_i \in \mathbb{R} \;|\; i \in V)$ and 
$\vct{g} = (g_e \in \mathbb{R} \;|\; e \in E)$, such that they are congruent to $\vct{y}$ and 
$\vct{m}$, we find a cost-minimal 
segmentation by minimizing
\vspace{-0.5mm}
\begin{equation}
\left<\vct{y},\vct{f}\right> + \left<\vct{m},\vct{g}\right>
\text{.}
\vspace{-0.5mm}
\end{equation}
However, not every assignment of the
indicators $\vct{y}$ and $\vct{m}$ results in a valid segmentation. We ensure consistency with the introduction of three types of constraints:
\emph{overlap constraints} ensure that no overlapping candidates are selected at the same time,
\emph{incidence constraints} force incident candidates of selected
adjacency edges to be selected as well, and
\emph{path constraints} state that for every adjacent pair of candidates that are not merged, there is no path of selected adjacency edges connecting them indirectly.
See \reffig{fig:cmc:overview:invalid} for
an illustration of these constraints. More formally, we require:
\begin{align}
  {\sum_{i\in C} y_i}
  &{\leq}
  {1}
  &{\forall C \in \mathcal{C}}
  \label{eq:cmc:ilp:overlap}
  \\
  {2m_{(i,j)} - y_i - y_j}
  &{\leq}
  {0}
  &{\forall (i,j) \in E}
  \label{eq:cmc:ilp:selection}
  \\
  {\sum_{e \in P} m_e - m_{(i,j)}}
  &{\leq}
  {|P| - 1}
  &{\forall (i,j) \in E, \forall P \in \mathcal{P}_{(i,j)}}
  \label{eq:cmc:ilp:path}
  \text{.}
\end{align}
For the overlap constraints in \refeq{eq:cmc:ilp:overlap}, $\mathcal{C} \subset 
2^V$ denotes the set of all \emph{conflict cliques}, \ie, a set of candidates 
that are mutually overlapping.
For tree-shaped candidate subset graphs like those used here, 
the set $\mathcal{C}$ simply contains all candidates of all paths from a 
leave node to the root node. In the example in 
\reffig{fig:cmc:overview:cmc}, $\mathcal{C} = \{ \{ a, e, g\}, \{b, e, 
g\}, \{c, f, g\}, \{d, f, g\} \}$.
The incidence constraints in \refeq{eq:cmc:ilp:selection} force the indicators $y_i$ and $y_j$ to be selected, if $m_{(i,j)}$ is selected.
Finally, the path constraints in \refeq{eq:cmc:ilp:path} ensure that if 
an adjacency edge was not selected (\ie, $m_{(i,j)} = 0$), there is no path of 
selected adjacency edges connecting them otherwise (\ie, the sum of selected 
edges along the path is strictly smaller than the length of the path). 
$\mathcal{P}_{(i,j)} \subset 2^E$ denotes the set of all paths on adjacency 
edges connecting candidates $i$ and $j$. Since there are in general 
exponentially many paths connecting two candidates in a CRAG, we do to not 
add these constraints initially. Following a cutting plane strategy, 
we solve an ILP without those constraints, and add violated constraints as
needed and resolve until a consistent solution is found.

\section{Experiments and Results}
\label{sec:results}

\begin{figure}[t]
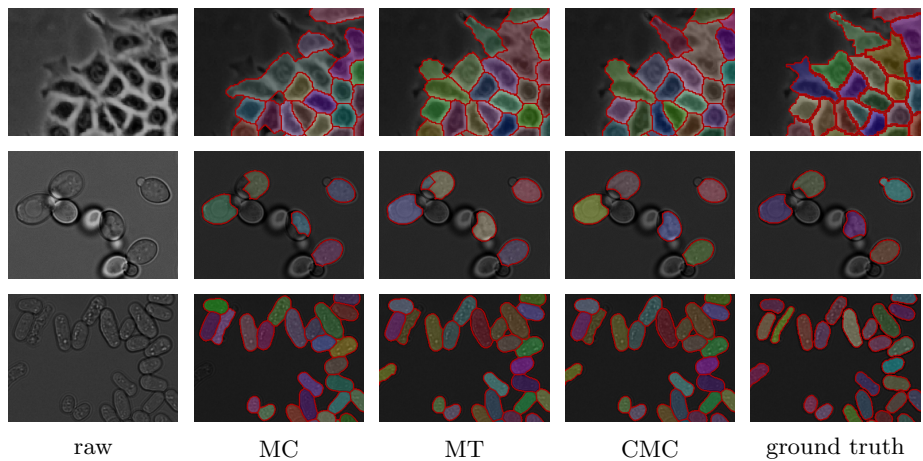

  \centerline{\includetikz{figures/results/samples}}
  \vspace{3mm}
  \caption{Samples of the segmentations obtained on datasets A, B, and C (from top to bottom), using the Multi-Cut formulation (MC)~\cite{Andres2012}, the merge-tree segmentation (MT)~\cite{Funke2015a}, and our model (CMC).}
  \label{fig:results:samples}
  \vspace{-2mm}
\end{figure}

\begin{table}[t]
\def\scale#1#2{\tikz[scale=#1,transform shape]{\node{#2};}}

\centerline{
\tikz[every node/.style={scale=0.925,transform shape}]{
\matrix[ampersand replacement=\&,column 2/.style={anchor=west}]{
\node{%
  \gdef\dataset{hela}%
  \gdef\datasetname{Dataset A}%
  \pgfplotstableread{figures/results/data/\dataset_mc}\mc%
\pgfplotstableread{figures/results/data/\dataset_tobasl}\tobasl%
\pgfplotstableread{figures/results/data/\dataset_cmc}\cmc%
\pgfplotstableread{figures/results/data/\dataset_tobasl_best-effort}\tobaslbe%
\pgfplotstableread{figures/results/data/\dataset_cmc_best-effort}\cmcbe%
\def\record#1#2#3#4#5#6#7{%
\pgfplotstabletypeset[%
  begin table={},%
  end table={},%
  skip coltypes,%
  columns={%
      VOI_SPLIT,%
      VOI_MERGE,%
      VOI,%
      DO_PRE,%
      DO_REC,%
      DO_FS
  },%
  columns/VOI_SPLIT/.style={fixed, fixed zerofill, precision=3},%
  columns/VOI_MERGE/.style={fixed, fixed zerofill, precision=3},%
  columns/VOI/.style={fixed, fixed zerofill, precision=3},%
  columns/RAND/.style={fixed, fixed zerofill, precision=3},%
  columns/DO_PRE/.style={fixed, fixed zerofill, precision=3},%
  columns/DO_REC/.style={fixed, fixed zerofill, precision=3},%
  columns/DO_FS/.style={fixed, fixed zerofill, precision=3},%
  columns/DO_MEAN_M1/.style={fixed, fixed zerofill, precision=3},%
  columns/DO_STD_M1/.style={fixed, fixed zerofill, precision=3},%
  columns/DO_MEAN_M2/.style={fixed, fixed zerofill, precision=3},%
  columns/DO_STD_M2/.style={fixed, fixed zerofill, precision=3},%
  columns/DO_MEAN_DICE/.style={fixed, fixed zerofill, precision=3},%
  columns/DO_STD_DICE/.style={fixed, fixed zerofill, precision=3},%
  every head row/.style={output empty row},%
  every row 0 column #3/.style={%
    postproc cell content/.append style={%
      /pgfplots/table/@cell content/.add={$\bf}{$}%
    }%
  },%
  every row 0 column #4/.style={%
    postproc cell content/.append style={%
      /pgfplots/table/@cell content/.add={$\bf}{$}%
    }%
  },%
  every row 0 column #5/.style={%
    postproc cell content/.append style={%
      /pgfplots/table/@cell content/.add={$\bf}{$}%
    }%
  },%
  every row 0 column #6/.style={%
    postproc cell content/.append style={%
      /pgfplots/table/@cell content/.add={$\bf}{$}%
    }%
  },%
  every row 0 column #7/.style={%
    postproc cell content/.append style={%
      /pgfplots/table/@cell content/.add={$\bf}{$}%
    }%
  },%
  write to macro=#2,%
  debug=true,%
  typeset=false]#1%
}%
\def\nohighlight{100}%
\def\pombedataset{pombe}%
\def\heladataset{hela}%
\ifx\heladataset\dataset%
  \record\mc\mcrow\nohighlight\nohighlight\nohighlight\nohighlight\nohighlight%
  \record\tobasl\tobaslrow{6}\nohighlight\nohighlight\nohighlight\nohighlight%
  \record\cmc\cmcrow{2}{5}{6}{7}\nohighlight%
  \record\tobaslbe\tobaslberow\nohighlight\nohighlight\nohighlight\nohighlight\nohighlight%
  \record\cmcbe\cmcberow{2}{5}\nohighlight\nohighlight\nohighlight%
\else%
\ifx\pombedataset\dataset%
  \record\mc\mcrow{6}{7}\nohighlight\nohighlight\nohighlight%
  \record\tobasl\tobaslrow{2}{5}\nohighlight\nohighlight\nohighlight%
  \record\cmc\cmcrow{2}{5}\nohighlight\nohighlight\nohighlight%
  \record\tobaslbe\tobaslberow\nohighlight\nohighlight\nohighlight\nohighlight\nohighlight%
  \record\cmcbe\cmcberow{2}{5}\nohighlight\nohighlight\nohighlight%
\else%
  \record\mc\mcrow{6}{7}\nohighlight\nohighlight\nohighlight%
  \record\tobasl\tobaslrow\nohighlight\nohighlight\nohighlight\nohighlight\nohighlight%
  \record\cmc\cmcrow{2}{5}\nohighlight\nohighlight\nohighlight%
  \record\tobaslbe\tobaslberow\nohighlight\nohighlight\nohighlight\nohighlight\nohighlight%
  \record\cmcbe\cmcberow{2}{5}\nohighlight\nohighlight\nohighlight%
\fi%
\fi%
\begin{tabular}{|r|rrr|rrr|}%
\hline%
&%
\multicolumn{3}{c|}{VOI}%
&%
\multicolumn{3}{c|}{DS}%
\\%
\hline%
&%
split &%
merge &%
total &%
prec. &%
rec. &%
F-sc.%
\\%
\hline%
MC &%
\mcrow%
MT &%
\tobaslrow%
CMC &%
\cmcrow%
\hline%
MT$^*$ &%
\tobaslberow%
CMC$^*$ &%
\cmcberow%
\hline%
\rowcolor{black!20!white}%
\multicolumn{7}{|c|}{\textsc{\datasetname}} \\%
\hline%
\end{tabular}%
};
\&
\node{%
  \gdef\dataset{yeast}%
  \gdef\datasetname{Dataset B}%
};
\\
\node{%
  \gdef\dataset{pombe}%
  \gdef\datasetname{Dataset C}%
};
\&
\node{%
  \gdef\dataset{snemi_small-sp}%
  \gdef\datasetname{Dataset D}%
  \pgfplotstableread{figures/results/data/\dataset_mc}\mc%
\pgfplotstableread{figures/results/data/\dataset_am}\am%
\pgfplotstableread{figures/results/data/\dataset_cmc}\cmc%
\pgfplotstableread{figures/results/data/\dataset_am_best-effort}\ambe%
\pgfplotstableread{figures/results/data/\dataset_cmc_best-effort}\cmcbe%
\def\record#1#2#3#4#5#6#7{%
\pgfplotstabletypeset[%
  begin table={},%
  end table={},%
  skip coltypes,%
  columns={%
      VOI_SPLIT,%
      VOI_MERGE,%
      VOI,%
      RAND
      },%
  columns/VOI_SPLIT/.style={fixed, fixed zerofill, precision=3},%
  columns/VOI_MERGE/.style={fixed, fixed zerofill, precision=3},%
  columns/VOI/.style={fixed, fixed zerofill, precision=3},%
  columns/RAND/.style={fixed, fixed zerofill, precision=3},%
  every head row/.style={output empty row},%
  every row 0 column #3/.style={%
    postproc cell content/.append style={%
      /pgfplots/table/@cell content/.add={$\bf}{$}%
    }%
  },%
  every row 0 column #4/.style={%
    postproc cell content/.append style={%
      /pgfplots/table/@cell content/.add={$\bf}{$}%
    }%
  },%
  every row 0 column #5/.style={%
    postproc cell content/.append style={%
      /pgfplots/table/@cell content/.add={$\bf}{$}%
    }%
  },%
  every row 0 column #6/.style={%
    postproc cell content/.append style={%
      /pgfplots/table/@cell content/.add={$\bf}{$}%
    }%
  },%
  every row 0 column #7/.style={%
    postproc cell content/.append style={%
      /pgfplots/table/@cell content/.add={$\bf}{$}%
    }%
  },%
  write to macro=#2,%
  debug=true,%
  typeset=false]#1%
}%
\def\nohighlight{100}%
\def\snemidataset{snemi_small-sp}%
\ifx\snemidataset\dataset%
  \record\mc\mcrow\nohighlight\nohighlight\nohighlight\nohighlight\nohighlight%
  \record\am\amrow\nohighlight\nohighlight\nohighlight\nohighlight\nohighlight%
  \record\cmc\cmcrow{2}{3}\nohighlight\nohighlight\nohighlight%
  \record\ambe\amberow\nohighlight\nohighlight\nohighlight\nohighlight\nohighlight%
  \record\cmcbe\cmcberow{2}{3}\nohighlight\nohighlight\nohighlight%
\fi%
\begin{tabular}{|r|rrr|r|}%
\hline%
&%
\multicolumn{3}{c|}{VOI} &%
\multicolumn{1}{c|}{RAND} \\%
\hline%
&%
split &%
merge &%
total &%
\\%
\hline%
AM &%
\amrow%
MC &%
\mcrow%
CMC &%
\cmcrow%
\hline%
AM$^*$ &%
\amberow%
CMC$^*$ &%
\cmcberow
\hline%
\rowcolor{black!20!white}%
\multicolumn{5}{|c|}{\textsc{\datasetname}} \\%
\hline%
\end{tabular}%
};
\\
}; 
} 
} 

\caption{Segmentation and detection results on datasets A, B, and C (2D light microscopy) and segmentation results on dataset D (3D electron microscopy). The best values of the used measures are highlighted: variation of information (VOI), detection score (DS)~\cite{Funke2015a}, and Rand index (RAND). We compare the standard Multi-Cut formulation (MC)~\cite{Andres2012} and the merge-tree segmentation (MT)~\cite{Funke2015a} against our model (CMC) on datasets A, B, and C. For the neuron reconstruction in dataset D, we compare against the assignment model (AM)~\cite{Funke2012} and the Multi-Cut (MC). Entries with $^*$ have been found by matching the ground truth as closely as possible. They show the best achievable result given the initial superpixels on the respective dataset and thus show the performance limit of each method. Note that MT and AM have little room for improvement (using better features, for instance), whereas the CMC could benefit a lot from better features or different learning methods.}
\vspace{-9mm}
\label{tab:results:abcd}
\end{table}

\noindent
{\bf Datasets.}
We validate our method on four datasets (see examples in \reffig{fig:results:samples}) which differ greatly in 
image resolution, microscopy modality, and cell type (shape, appearance and 
size):
(A) phase contrast images of cervical cancer Hela cells~\cite{Arteta12}
%
(B) bright field images of in-focus Diploid yeast cells~\cite{Zhang2014}
%
(C) bright field images of Fission yeast cells~\cite{Peng2013}, and
(D) 3D serial section electron microscopy volume of neural tissue~\cite{Arganda-Carreras2013}.


\noindent
{\bf Merge-Tree Generation.}
Our model requires us to generate a merge-tree to build the CRAG.
For datasets A, B, and C, we ran a seeded watershed algorithm to obtain an initial oversegmentation on the pixel-wise boundary predictions from~\cite{Funke2015a}. We obtained a merge-tree by iteratively merging neighboring superpixels with minimal merge score. The merge score we used is the product of the smaller region's size and the median intensity of the boundary pixels separating the neighbors. From this merge-tree, we included all candidates that are the result of 5 or less merges in the CRAG, and added adjacency edges between each pair of touching candidates (considering a 4-neighborhood) across all levels.
For dataset D, we generated merge-trees in a similar fashion for each 2D image of the stack individually, using the boundary predictions from~\cite{Arganda-Carreras2013}. In the CRAG, we only included the leaf nodes of the merge-tree and the largest candidates below a merge score that still favors oversegmentation.

\noindent
{\bf Features.}
For all experiments, we extracted the same features for each candidate: size, 
circularity, eccentricity, and a contour angle histogram 
with 16 bins. In addition to that, we extracted intensity features (sum, mean, variance, skewness, kurtosis, histogram with 20 bins, 
7 histogram quantiles) from the raw and boundary 
prediction images for the whole candidate and for the contour pixels.
For edge features, we used the size of the contact area between neighboring candidates, as well as mean, variance, and skewness of the intensities across the boundary pixels separating the neighbors. In addition to that, we added the absolute difference, min, max, and sum of each feature of the two adjacent candidates.

\noindent
{\bf Training.}
The training consists of finding suitable costs for candidates and edges, given a training dataset with a CRAG and a ground truth image or volume.
To learn the costs, we trained a random forest classifier by first generating a consistent CMC solution that is closest to the ground truth.
For that, we assign each leaf node to the ground truth region with largest overlap.
We then selected all candidates that cover leaf nodes with the same label as positive candidate instances.
Similarly, we selected edges linking selected candidates with the same label as positive edge instances.
All other candidates or edges were considered negative instances.
Note that this training setup includes edges spanning different levels of the CRAG, and thus allows the classifier to relate features of candidates of different sizes.
For datasets A, B, and C, we only trained the edge costs as described, and learned the node costs as suggested in~\cite{Funke2015a}.

\noindent
{\bf Comparison.}
We compare the performance of our model on datasets A, B, and C against a Multi-Cut (MC)~\cite{Andres2012} and the merge-tree (MT) method described in~\cite{Funke2015a}. For both, we use exactly the same candidates (for MT), edges of leaf nodes (for MC), features, and learning method as described above.
On dataset D, we compare against a Multi-Cut (MC) and the assignment model (AM) proposed in~\cite{Funke2012}, which is a variation of a merge-tree method. Again, we used the same candidate hierarchy for AM, the same adjacency edges for MC, and the same learning method as used for CMC.
We trained each method on the same training subset of each dataset, and report results on the remaining images or volume.

\noindent
{\bf Evaluation.}
For datasets A, B, and C, we report two measures, as in~\cite{Funke2015a}: variation 
of information (VOI) to measure segmentation accuracy and detection score (DS) to measure the detection accuracy (see~\reftab{tab:results:abcd}).
  The Candidate Multi-Cut improves accuracy on two of the three datasets, and produces the exact same result as MT on dataset B. In the same table, we also report the results of the best achievable segmentation of each method, denoted as MT$^*$ and CMC$^*$. We found these segmentations in the same way we generated the training samples as described above. It can clearly be seen that CMC has a higher expressiveness, although it uses the same initial superpixels as MT.
For dataset D, we report VOI and the Rand index (RAND) to measure segmentation accuracy (see~\reftab{tab:results:abcd}). The Candidate Multi-Cut improves accuracy compared to both MC and AM. Again, we report the best achievable segmentation as AM$^*$ and CMC$^*$ and find CMC to be more expressive than AM.

\section{Conclusions}

We presented a generalization of two successful segmentation methods for biomedical images and demonstrated that our model combines the advantages of both methods.
This is a worthy contribution on its own as it broadens the range of applications that a single method can be used for.
We find this especially remarkable as our method does not introduce new hyperparameters, and thus its advantages come basically ``for free''.
On top of that, we could report a modest improvement in segmentation and detection accuracy which shows that our generalized model is more than competitive with its specialized variants.
Considering that our method has a higher expressiveness compared to merge-tree methods and larger context than Multi-Cut methods, we believe that further improvements are possible by learning node and edge costs in a more principled way.
We believe that this, together with the design or learning of features that are more discriminative, is a promising direction for further research.

\bibliographystyle{plain}
\bibliography{library}

\end{document}